\documentclass[letterpaper, 10 pt, conference]{ieeeconf}  

\usepackage{stfloats}
\usepackage{times}
\usepackage{epsfig}
\usepackage{graphicx}
\usepackage{amsmath}
\usepackage{amssymb}
\usepackage{algorithm}
\usepackage[noend]{algpseudocode}
\usepackage{subfigure}
\usepackage{hhline}
\usepackage{color, colortbl}

\IEEEoverridecommandlockouts                              

\title{\LARGE \bf
Explicit Domain Adaptation with Loosely Coupled Samples
}

\author{Oliver Scheel$^{1, 2}$, Loren Schwarz$^{1}$, Nassir Navab$^{2}$, Federico Tombari$^{2, 3}$
\thanks{$^{1}$BMW Group, M\"{u}nchen, Germany
        {\tt\small first.lastname@bmw.de}}%
 \thanks{$^{2}$Faculty of Computer Science, Technische Universit\"{a}t M\"{u}nchen,
Garching bei M\"{u}nchen, Germany}%
\thanks{$^{3}$Google Inc., Switzerland}%
}

\definecolor{tblue}{rgb}{0,0, 1}
\definecolor{tred}{rgb}{1,0, 0}
\definecolor{tgreen}{rgb}{0.2,0.58, 0.2}
\definecolor{tyellow}{rgb}{0.78,0.78, 0.16}

\begin{document}

\setlength{\textfloatsep}{5pt plus 1.0pt minus 2.3pt}

\maketitle
\thispagestyle{empty}
\pagestyle{empty}

\begin{abstract}
Transfer learning is an important field of machine learning in general, and particularly in the context of fully autonomous driving, which needs to be solved simultaneously for many different domains, such as changing weather conditions and country-specific driving behaviors. Traditional transfer learning methods often focus on image data and are black-box models.
In this work we propose a transfer learning framework, core of which is learning an explicit mapping between domains. Due to its interpretability, this is beneficial for safety-critical applications, like autonomous driving. We show its general applicability by considering image classification problems and then move on to time-series data, particularly predicting lane changes. In our evaluation we adapt a pre-trained model to a dataset exhibiting different driving and sensory characteristics.
\end{abstract}


\section{INTRODUCTION}
Transfer learning aims to apply models to domains different from those they were originally trained on - possibly even to different tasks - while leveraging existing knowledge.
This process, which is mastered exceptionally well by humans, is an interesting and important field of research, and an important step towards the concept of general artificial intelligence. Further, it mitigates some challenges of current deep learning algorithms, such as the need to collect and annotate huge amounts of data for each domain and task.

\begin{figure}[!t]
\centering
\includegraphics[scale=0.6]{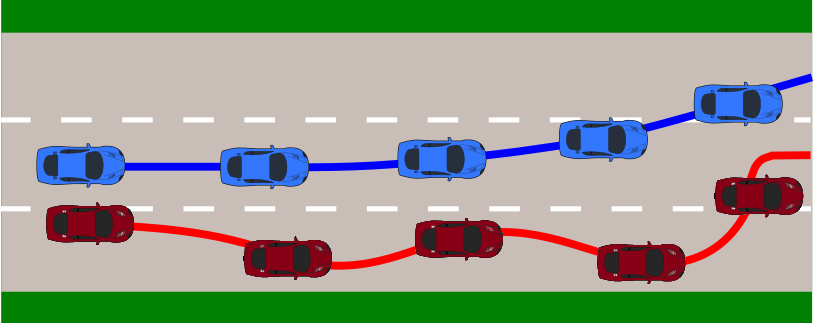}
\caption{Comparison of two lane changes in different domains. The blue car executes its lane change smoothly, while the red one exhibits a noisy driving style, causing many false predictions in models not exposed to this.}
\label{fig:teaser}
\end{figure}

In the field of autonomous driving, transfer learning also plays a crucial role, but is among the lesser explored topics. Here, many domains are conceivable and necessary, models need to be able to cope with a variety of complex environments and situations:
a camera system needs to produce reliable detections under varying weather and illumination conditions - a trajectory prediction model has to function for rural areas and crowded megacities, considering location- and region-specific driving behaviors (see Figure \ref{fig:teaser} as an example).
Other changes stem from within the vehicle itself: when sensor configurations or software versions change, do neural networks from previous versions have to be trained from scratch?
Furthermore, model transfer is strongly coupled with the problem of safety: having demonstrated that a model satisfies certain functional safety requirements, can we move to new domains so that these guarantees still hold?

This gives rise to the need for explainable transfer learning techniques. In this work we propose such a method, learning an explicit mapping between domains in the form of a transformation matrix. Related work  targets image-to-image translation and style transfer, often using adversarial approaches \cite{8237506}, which are not interpretable and further focus only on image data.
Although explicit domain adaptation algorithms have been proposed, often they are specific to certain tasks and models \cite{aswolinskiy2017unsupervised, paassen2016linear}, whereas we introduce a more general framework with a novel correspondence loss.

Core of our proposed method is learning a mapping, with which samples can be converted between domains. Advantages of such an explicit transformation are several, e.g. the previously mentioned interpretability but also the possibility of integrating domain knowledge 
by pre-training or initialization.
We test our approach on different toy tasks, covering image and sequence problems, and on the real-world problem of predicting lane changes. For this we tackle the challenging transfer task of varying sensor-setup and driving styles, achieving promising results.

\section{RELATED WORK}
In literature, usage of terms like transfer learning and domain adaptation is sometimes inconsistent. Here we follow the notation of \cite{5288526}, denoting with transfer learning the general concept and with domain adaptation the branch of shifting probability distributions over domains to make them more similar. 
Thus our approach mostly falls into the latter category, but also exhibits shares of other branches.

One transfer learning technique is fine-tuning of pre-trained networks, for which lower levels of the network are frozen, while only layers close to the output are trained on the new domain or task. This often outperforms training from scratch, due to the amount of information already stored in the network, and leads to state-of-the-art algorithms, even for totally different tasks \cite{vqa, antol2015vqa}.
Another common approach is using adversarial methods to generate data resembling the desired distribution, often applied in the context of image translation and for creating synthetic training data \cite{NIPS2016_6544, 8099501, synth}. 
Few- or one-shot learning aims to understand new concepts with only few annotated samples \cite{koch2015siamese, finn2017model}.
Multi-task learning specializes in leveraging common information for solving different tasks, often with a shared base-architecture \cite{mtl}.


Yim et al. extend the concept of knowledge distillation \cite{kd} to include transfer learning applications with a focus on network compression \cite{8100237}. 
Core idea of their approach is the matching of feature activations over different layers and domains, which is commonly seen in other works as well \cite{Luo:2017:LEL:3294771.3294787, bswf}.

Domain adaptation methods have been frequently applied in the field of image translation, e.g. for transferring image styles or creating synthetic training images in different domains \cite{8099501, synth}.
Zhu et al. achieve fascinating results for unpaired image-to-image translation, using two Generative Adversarial Networks (GANs) and a cycle consistency loss \cite{8237506}. In contrast to this, in our method we make use of existing (loosely) coupled training samples, and calculate an explicit transformation matrix $\mathbf{T}$. This simplifies understanding of the process, and is closer to the underlying idea, transforming samples between domains instead of hallucinating them. 
Since $\mathbf{T}$ nearly always is invertible, we are given the cyclic conversion ``for free'' and save parameters.
Isola et al. employ a similar principle as \cite{8237506}, except using exact correspondence pairs $(x, y)$ and only one generator and discriminator \cite{DBLP:journals/corr/IsolaZZE16}. The generator is conditioned on input $x$ from domain A, and tries to create the corresponding image $y$ from domain B, s.t. the discriminator cannot distinguish from real samples of these domains.
Li et al. employ linear transformation matrices for the problem of style transfer, achieving state-of-the-art results \cite{li2018learning}. In comparison to their work, which calculates a style loss at different layers in the network, we use a correspondence loss in the actual domains, making transformations better understandable and also enabling usage in shallower architectures, like Recurrent Neural Networks (RNNs). 
Jaderberg et al. address the weakness of classical Convolutional Neural Networks (CNNs) and introduce Spatial Transformer Networks (STNs) to deal with transformed images \cite{jaderberg2015spatial}. Their goal neither is transfer learning and they lack the correspondence loss as we introduce it, 
still their idea is similar. 

Duan et al. introduce transformation matrices to project data of different domains in a common space, and include these in Support Vector Machines (SVMs) \cite{Duan:2012:LAF:3042573.3042661}.
Paa{\ss}en et al. also extend a common optimization concept with an explicit transformation matrix $\mathbf{H}$ 
\cite{Paaen2016LinearST}. While their transformation $\mathbf{H}$ is global, we calculate a transformation $\mathbf{T}_x$ for every datapoint $x$, which allows for greater flexibility. Paa{\ss}en only include $\mathbf{H}$ in the loss formulation, thus using a very similar loss to \cite{jaderberg2015spatial}.
Aswolinskyi et al. extend \cite{Duan:2012:LAF:3042573.3042661} to unsupervised applications, learning a predictive model to generate synthetic data for domain $B$ in an autoregressive fashion \cite{aswolinskiy2017unsupervised}. 

Sun et al. introduce a preprocessing procedure, dubbed CORAL, to address domain shift and leverage transfer between domains \cite{10.5555/3016100.3016186}. They propose, in addition to the common standardization step of modifying data to possess 0 mean and unit variance, to further minimize distances in second-order statistics (covariance) between both analyzed datasets.

Triplet loss is a common method for defining similarities of data points \cite{7298682}, and thus related to our interpretation of correspondence. For more details we refer to Section \ref{sec:corrs}.
\section{PROBLEM DEFINITION}
In all our experiments we use a source domain $A$ and target domain $B$, on which we solve the same task. We have a (complex) model $\mathbf{M}$ trained on $A$ and aim to leverage this knowledge to domain $B$. We assume, that data for $A$ is plenty and sufficient, while for $B$ it is more rare, which is a realistic application scenario. To simulate this and further analyze the effect of data quantity on transfer success, we limit domain $B$ to $b$ data samples for varying $b$.

In unison with the general idea of transfer learning, it is not desired to simply train $\mathbf{M}$ from scratch on $B$, as this has several disadvantages, in general, but in particular also for the field of autonomous driving: it is not desirable to train and store full, complex models for each domain (e.g. for each existing country), due to time and resource constraints. Further, any 
behavioral analysis of $\mathbf{M}$, e.g. for safety verification reasons - like running a simulation, had to be done anew. Additionally, for small $b$ our experiments show bad results when training full models, among others due to the disproportion of available training data and model parameters. This was also noted by \cite{8100237}.
Due to this, we aim at leveraging and incorporating existing knowledge, and to build on models pre-trained on $A$.

\subsection{Corresponding Samples}
\label{sec:corrs}
Core of our proposed algorithm is a novel correspondence loss, for which we need corresponding samples from both domains. There are different ``degrees'' of correspondence though, and we briefly discuss our needed one. \cite{8237506} consider completely unpaired samples. In their method, random samples from domains $A$ and $B$ are drawn for training, and 
a generator learns to generate samples from the other domain, indistinguishable for a discriminator. Conversely, \cite{DBLP:journals/corr/LongSD14} fine-tune classification models for the task of image segmentation. When considering the problem of mapping images to semantic segmentation maps as domain adaptation task, images $x$ in the original domain  are paired with the corresponding segmentation $x'$. 
This is (depending on the labelling quality) nearly an exact correspondence, there exists a simple mapping $f$ s.t. $f(x) = x'$. In \cite{8237506}, this mapping neither is simple nor deterministic. 

For our application scenario, we relax this assumption to a rough correspondence, namely  $f(x) \approx x'$, guiding our model towards the correct solution by providing $n$ correspondence samples $(x, x_1'), \ldots, (x, x_n')$, s.t. the learned average mapping provides a good fit - thus the name ``loosely coupled''. For images to semantic views, this corresponds to contrasting the original image of a scene with $n$ semantic segmentation maps, which are slightly modified, e.g. differ in the number of parked cars on a street. 

For our experiments on MNIST, an image is assigned $n$ images from the other domain with equal label.
For the problem of predicting lane changes, sequences are aligned based on labels, as well: Correspondences of follow-only sequences are follow-only sequences of the other domain. If a lane change is present, correspondences are lane changes of the other domain, aligned s.t. the execution times of maneuvers match as closely as possible. 
Throughout all our experiments, we use $n = 5$. This decision is based on empirical evidence of good performance, and further denotes a realistic number of available correspondence pairs.

\begin{figure}[!t]
\centering
\includegraphics[scale=1]{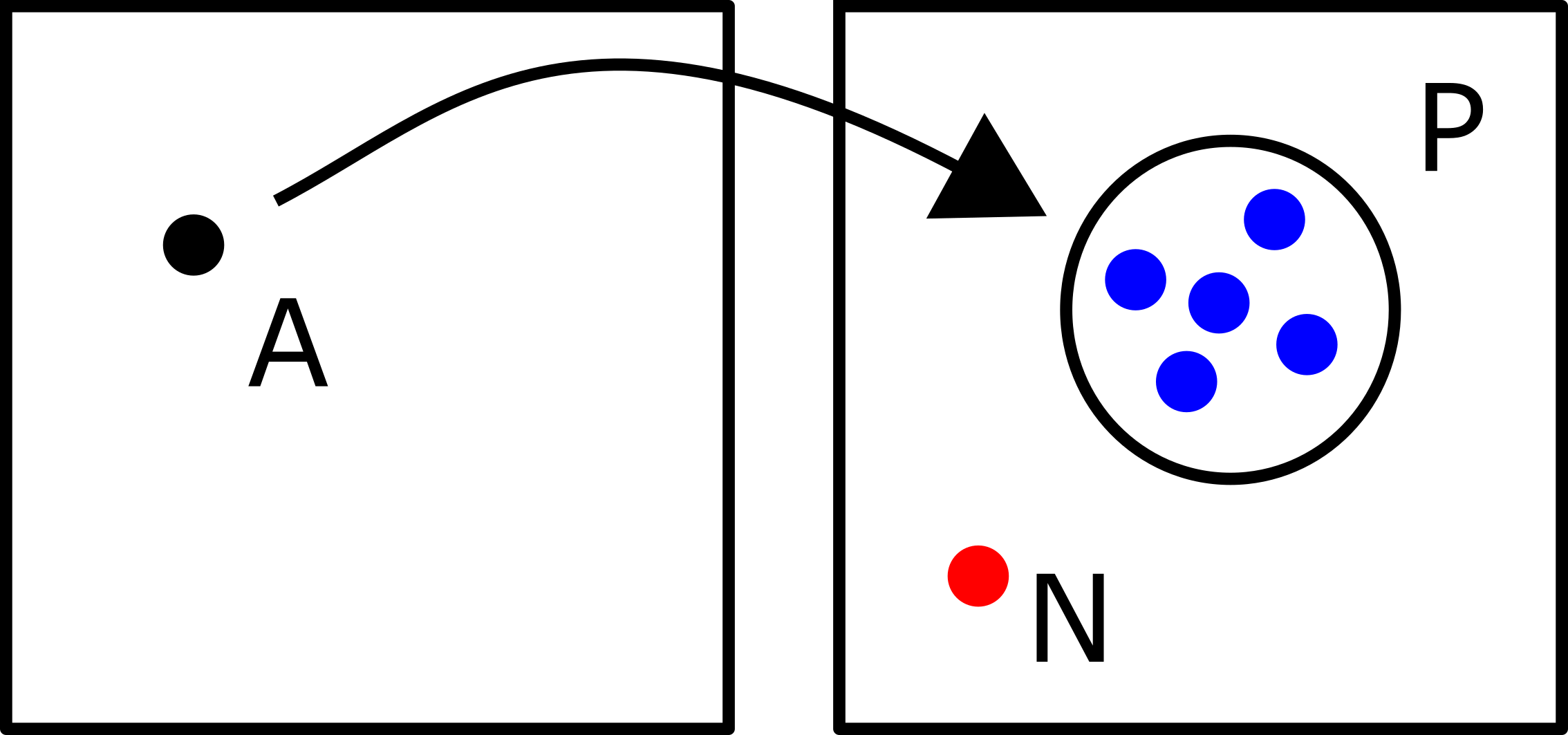}
\caption{Depiction of our used correspondence mapping $f$ and its relation to triplet loss. For the triplet loss, an anchor point (A) is assigned a positive sample (\textcolor{tblue}{P}) and contrasted to a negative one (\textcolor{tred}{N}). In our case, $f$, and thus the model trained with it, is a generative way of converting anchor point A from one domain to another (ideally, resulting in the mean of the positive correspondence points).}
\label{fig:samples}
\end{figure}

To conclude this section, note the relation of our interpretation of correspondence to triplet loss \cite{7298682}. Triplet loss is often used in applications, where neural networks are supposed to learn metric distances between samples. This way, among others the problem of object re-identification can be tackled. The triplet loss of an anchor point $a$ and a negative ($n$) and positive sample ($p$) is defined as $L = max(0, d(a, p) - d(a, n) + m)$, in which $m$ describes a minimal margin. Thus, a model optimized using this loss is trained to map samples into a latent space, s.t. the distance $d$ between aligned samples is small, and large for unrelated samples. Our correspondence samples $(x, x_1'), \ldots, (x, x_n')$ can be understood as pairs of anchor point $x$ and positive samples $x_i'$, and no negative samples are given. While models trained with triplet loss offer a discriminative procedure of deciding which inputs are close together, our proposed framework can be understood as a generative model for sampling such positive points (see Figure \ref{fig:samples} for a visualization).

\section{MODEL}
Core of our proposed framework is a neural network we call Converter ($\mathbf{C}$), which is responsible for converting samples from one domain to another. $\mathbf{C}$ is prepended before the actual network trained on $A$, and calculates a transformation matrix $\mathbf{T}_x$ for each input $x \in B$. $x$ is then multiplied by $\mathbf{T}_x$, which equals applying the learned mapping into domain $A$, s.t. the resulting value resembles the corresponding input in domain $A$.

Formally, let $x$ be a sample of domain $B$, $x\in \mathbb{R}^d$, $d \in \mathbb{N}$, with samples of $A$ having the same dimensionality, and let $\mathbf{x}$ be the homogeneous representation of $x$, i.e. $\mathbf{x} = \begin{pmatrix}x&1 \end{pmatrix}^\top$. Then $\mathbf{C}$ defines the mapping
\begin{equation}
\mathbf{C}: B \rightarrow \mathbb{R}^{(d+1) \times (d+1)}, \ x \mapsto \mathbf{T}_x 
\end{equation}
s.t.
\begin{equation}
\mathbf{T}_x \mathbf{x} = \mathbf{x}'    
\end{equation}
and $x'$ is a good representation of $x$ in $A$. 
Assume $\mathbf{M}$ is our full model trained on $A$, then let $\mathbf{L}$ denote the last $l$ layers of $\mathbf{M}$ connected to the output.
$\mathbf{C}$ usually has fewer parameters than $\mathbf{M}$, as it is intended for a simpler purpose, solving part of the original task, which helps saving resources.

\begin{figure}[!t]
\centering
\includegraphics[scale=0.3]{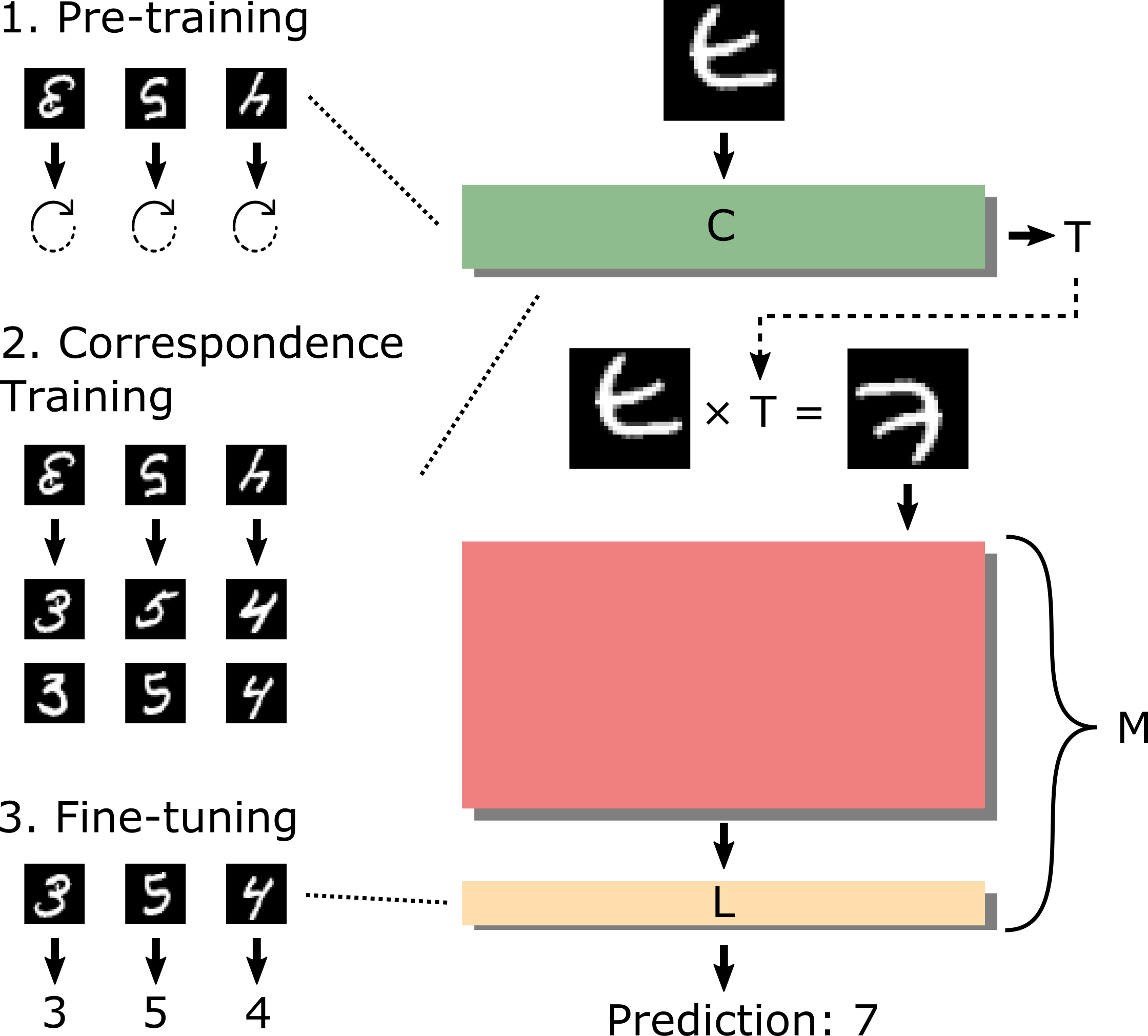}
\caption{Graphical overview of our framework adapting a model trained on standard MNIST to rotated images. Steps 1 and 2 consist of training the Converter $\mathbf{C}$, first pre-training it with (an expected) rotation, and then with $n$ correspondence pairs for each sample (here $n = 2$). In Step 3, the last layers $\mathbf{L}$ of $\textbf{M}$ are fine-tuned on the new dataset, while the complex part of $\textbf{M}$ is frozen (drawn in red).}
\label{fig:model}
\end{figure}

Training of our framework consists of three steps, each of which is optional (e.g., it can also be used without corresponding samples):
\begin{enumerate} 
\item \textbf{Pre-training}: Initialize $\mathbf{C}$ with an educated guess of possible transformation matrices. For this, train $\mathbf{C}$ with samples $x \in B$, using some initial transformation matrix $\mathbf{\Tilde{T}}_x$ as ground truth and for optimization the element-wise L2-loss $\mathfrak{L}_P(x)$ (which equals the Frobenius norm in matrix space):
\begin{equation}
\mathfrak{L}_P(x) = \vert \mathbf{C}(x) - \mathbf{\Tilde{T}}_x \vert_F
\end{equation}
\item \textbf{Correspondence training}: Train $\mathbf{C}$ with the correspondence pairs $(x, x_1'), \ldots (x, x_n')$ for $x \in B$. Use a suited loss function $\mathfrak{L}_C(x)$ operable in the domain spaces, e.g. the L2-loss between converted and actual samples:
\begin{equation}
    \mathfrak{L}_C(x) = \frac{1}{n}\sum_{i=1}^{n} \vert \mathbf{C}(x) \mathbf{x} - \mathbf{x_i'}\vert_2
\end{equation} where again $\mathbf{x_i'}$ is the homogeneous version of $x_i'$.
\item \textbf{Fine-tuning}: For this, we distinguish two train modes: In mode 0, we only retrain $\mathbf{L}$ with the actual task loss (e.g. a cross-entropy loss for classification). In mode 1, we retrain $\mathbf{L}$ and $\mathbf{C}$, accumulating task loss and correspondence loss $\mathfrak{L}_C$.
\end{enumerate}
Step 1 enables incorporating existing domain knowledge into the framework, as often one might have a rough estimate of how the resulting transformation could look like (e.g. a conversion between Radar and Lidar points, variation of the angle of mounted cameras ...). If such prior information is not available, it is recommendable to use $\mathbf{\tilde{T}} = \mathbf{I}^{d+1}$. This way, if the converter module cannot deduce a meaningful intra-domain transformation, the method will still be as least as good as standard fine-tuning. In Step 2 the model makes use of the given corresponding samples, in order to further refine or learn the best transformation between samples from domains $A$ and $B$ in a data-driven fashion. Hopefully, by now the output of $\mathbf{C}$ resembles samples of domain $A$, s.t. the original model $\mathbf{M}$ already exhibits good results. As such transformation is not perfect though, in Step 3 the model is fine-tuned to further increase performance. Figure \ref{fig:model} shows a graphical overview of our framework.
\subsection{Simplifying Assumptions}
$\mathbf{T}$ represents a transformation matrix in the domain space, and as such can be restricted to only allow certain transformations, such as rotations, affine or projective transformations. The tighter the restriction, the more limited is the model, but this simultaneously makes learning the transformation easier. 
In addition to this, in the following we describe simplifying adaptations for image and sequential data. Note though that in any case it is possible to apply our framework without these, working directly in the full vector spaces of all images or sequences, but that these assumptions are simplifying approximations helpful in reducing complexity.
\subsubsection{Image Data}
Let $x$ be an image of size $w \times h$, thus $x \in \mathbb{R}^{w \times h}$. As image transformations are often structured (e.g. rotations, rectifying projections, ...), we apply transformations globally for the whole image, treating each pixel similarly. Instead of operating in the image space $\mathbb{R}^{w \times h}$, our transformations live in the space  $\mathbb{R}^{3 \times 3}$, representing the common homogeneous transformations for 2D-points. We utilize the module introduced and used in STNs to apply these transformations, although other approaches are possible.
\subsubsection{Sequential Data}
Main focus of this work is dealing with time-series data. For these we employ frame-based transformations, as well. Assume we examine sequences of length $l$ and feature size $f$, thus $x \in \mathbb{R}^{f \times l}$. Instead of learning accordingly dimensioned transformations, we consider each frame separately, s.t. $\mathbf{T} \in \mathbb{R}^{(f+1) \times (f+1)}$. To model the temporal context, in this case $\mathbf{C}$ is an RNN, which expects full sequences as inputs and outputs a sequence of transformation matrices. More specifically, we use RNNs consisting of Long-Short Term Memory Units (LSTMs) \cite{gers1999learning}.
We use the following shorthand notation for an LSTM cell:
\begin{equation*}
    (\mathbf{h}^t, \mathbf{\tilde{c}}^t) = \texttt{LSTM}(\mathbf{x}^t, \mathbf{h}^{t-1}, \mathbf{\tilde{c}}^{t-1})\\
\end{equation*}
where $\mathbf{x}^t$ is input at timestep $t$ and thus $t$-th frame of the sequence, $\mathbf{h}$ the hidden state and $\mathbf{\tilde{c}}$ the memory unit.
With weight matrix $\mathbf{W}$ and bias vector $\mathbf{b}$, the calculation of $\mathbf{T}$ in step $t$ is given by:
\begin{equation}
\label{eq:lstm}
\begin{split}
\mathbf{T}_x^t & = \texttt{softmax}(\mathbf{W}\cdot\mathbf{h}^t + \mathbf{b}).
\end{split}
\end{equation}

\section{DATASETS AND METRICS}
In this section we introduce used datasets and metrics for evaluation. We start with two toy problems to motivate and show functionality of our proposed model, and then move to the real-world problem of predicting lane changes. For each analyzed problem we describe the architecture of the ``base'' model $\mathbf{M}$, which is trained on domain $A$ and serves as starting point for the transfer task to domain $B$.
\subsection{Datasets}
\subsubsection{Transformed MNIST}
For simple motivation, we use images of the well-studied MNIST dataset \cite{lecun-mnisthandwrittendigit-2010}. 
While domain $A$ are the standard images, domain $B$ consists of images rotated by $180^{\circ}$ degrees. $\mathbf{M}$ is a simple CNN consisting of 3 convolutional and max-pooling layers, followed by one fully-connected layer for classification. For fine-tuning, $\mathbf{L}$ is this last layer.
\subsubsection{Toy Sequences}
We simulate lane change maneuvers, representing each solely by the target car's distance to the lane's center line - ranging from 0 (at left lane border) to 1 (at right lane border).  Domain $A$ consists of ``clean'' lane changes, in which drivers smoothly follow their lane until the maneuver, and then also execute this in an orderly fashion. In Domain $B$ we added noise to these trajectories, see Figure \ref{fig:teaser}. This could model the realistic scenario of transferring between country-specific driving styles - one can think of countries in which traffic is more strictly regulated and rule-abiding, while in some countries trajectories are more freely chosen. One can assume, that a model trained on $A$ and applied to $B$ exhibits more false predictions, due to its missing exposure to noise. 
Indeed, this is the case, as results in Section \ref{sec:results} show, even, when using high-level data as we do here. Naturally, the domain gap for problems concerning images or raw sensor data is greater, still, our results show a significant performance drop also for this abstracted data. $\mathbf{M}$ is similar as in the next paragraph, except using only one feature, namely distance to the lane's center line, and a smaller hidden size of 32. Further, also the problem definition including how ground truth labels are defined, is identical to the one in the next paragraph. In particular, this is a classification problem, in which each frame is given one of the labels ``follow lane'' or ``change lanes''.
\subsubsection{Lane Change Prediction}
For predicting lane changes, we closely follow our previous work \cite{lcp}, using the simple LSTM model as base model. This is an LSTM network of hidden size 64 with a single classification layer on top. For retraining, $\mathbf{L}$ is this classification layer. A multitude of features is available, 
for this work, though, for visualization purposes we mostly use the two features distance to lane's center line ($m$) and lateral velocity ($v$). These were proven to be the most relevant ones by \cite{DaimlerNB}. Each frame is given one of the labels $L$(eft), $F$(ollow) or $R$(ight). The model also predicts in each step, with a prediction of $L$ or $R$ indicating the algorithm's belief that a lane change to the respective side is imminent. In particular, all frames of the dataset are labelled $F$, but frames from 3s before a lane change until its execution (the time point of the vehicle crossing lane boundaries) are labelled with the respective direction. Shortly before this time period, and after a lane change, frames are assigned weight 0, to give models time to adjust and not penalize ``too early'' predictions (see \cite{lcp} for more details).
Due to the huge imbalance in the datasets (a majority of frames is labelled $F$), frames are weighted inversely proportional to their class' frequency. Further, frames labelled $L$ or $R$ are weighted exponentially more the closer the lane change (compare \cite{b4c, lcp}). 

Domain $A$ is a collection of lane changes collected by BMW in-series cars (introduced in \cite{lcp}), mainly in Germany. Domain $B$ is the publicly available NGSIM dataset \cite{ngsim}, which is a collection of lane changes recorded on highways in the United States. Next the different recording countries, different sensor-setups, e.g. with different sampling frequencies, were used. Although these differences are diminished by the used high-level representation, among others we interpolate between frames, causing different temporal characteristics. Further, recording for NGSIM was done in busy highway sections, predominantly during rush-hour and in sections with on- and off-ramps. This causes lots of ``unstable'' driving and many successful but also failed merging maneuvers. In contrast to this, recordings of the fleet data in general contain less dense traffic scenes. 
\subsection{Metrics}
For evaluating lane change problems, we follow the metrics introduced in \cite{lcp} and for simplicity and better comparison pick the three most relevant ones:
\begin{itemize}
    \item \textbf{Frequency}: number of times a maneuver is
predicted per ground truth event. We only consider lane-following periods for this, here this equals the number of false positives, i.e. the number of times a lane change is falsely predicted. An optimal model exhibits a \emph{Frequency} of 0.
    \item \textbf{Delay}: delay of lane change prediction w.r.t. ground truth, measured in seconds. A delay of 0s describes a prediction, which begins as soon as the ground truth label switches to lane change (and thus predicts the maneuver 3s ahead of its execution).
    \item \textbf{Miss}: number of lane changes completely missed.
\end{itemize}
To better compare algorithms, we further accumulate these into one score as follows: Model $\mathbf{M}$, which was trained on $A$ and tested on $B$, is used as baseline. For each subsequent algorithm, its percental increase or decrease w.r.t. this baseline in each metric is measured. The resulting score is the sum of these percentages, in which we weigh \emph{Frequency} by the relative number of frames labelled $F$, and \emph{Delay} and \emph{Miss} by the relative number of frames labelled $L$ or $R$. This way, we assign priorities to the maneuver classes according to the relative duration a driver experiences them for, and acknowledge the huge importance of \emph{Frequency}: A controller will and has to react to each false prediction, causing driver discomfort. Furthermore, a relative decrease in \emph{Delay} can be tolerated if in acceptable bounds, and the number of misses of our algorithms is so low, that small absolute changes here distort percentage values.
\section{RESULTS}
\label{sec:results}
We implement state-of-the-art baseline methods and compare our methods against these. We first introduce these in general, and in the respective sections describe any problem-specific adaptations, if existing.
\subsection{Baseline Methods}
\begin{itemize}
\item Fine-tuning: We fine-tune layers  $\mathbf{L}$ of model $\mathbf{M}$, which was trained on $A$ and tested on $B$. Note that \cite{8100237} compare against this as well, and perform marginally worse due to their focus on network compression, thus a relative comparison to their approach can be established.
\item CORAL \cite{10.5555/3016100.3016186}: We use the CORAL algorithm to minimize covariance distances between used domains $A$ and $B$, which have already been standardized. Afterwards, we fine-tune layers $\mathbf{L}$ of model $\mathbf{M}$ on the new domain.
\item pix2pix \cite{DBLP:journals/corr/IsolaZZE16}: To examine an implicit domain adaptation method, in contrast to our explicit one, we choose the conditional adversarial approach from the pix2pix framework.
\item Imp: To further analyze a possible trade-off between explicitness and model capabilities, this baseline equals our proposed framework, except $\mathbf{C}$ now does not output a transformation matrix $\mathbf{T}$, but directly samples from domain $B$. Due to this, Step 1 of our proposed training scheme is not possible and thus left out.
\item Mode 2: Further, we compare against \cite{Paaen2016LinearST} and STNs \cite{jaderberg2015spatial}. Note that a slight improvement of \cite{Paaen2016LinearST} (using a sample-specific transformation matrix instead of a global one) as well as STNs can be expressed by means of our framework: Both methods employ a transformation matrix, and fine-tune existing models adapting $\mathbf{C}$ and $\mathbf{L}$. They lack Steps 1 and 2, concerning pre-training and training the Converter (and lack our correspondence loss overall), but otherwise are identical to our algorithm with train mode 1, except only the classification loss is considered. We denote this small modification by Mode 2, and additionally add Step 1 with the same pre-training targets for fairness.
\item Mode 0: We use similar experiments as further ablation studies, fathoming how the correspondence loss affects performance, as this is one of our contributions and a core difference to \cite{Paaen2016LinearST} and STNs: Mode 0 denotes our framework stripped of Step 2 and using train mode 0, thus excluding the correspondence loss.
Both methods, Mode 2 and Mode 0, answer the question whether using such a correspondence loss is beneficial, or whether similar results can be obtained by just pre-training a converter well and eventually fine-tuning the model, including $\mathbf{C}$ or $\mathbf{L}$ only. 
\end{itemize}
\subsection{Rotated MNIST}
Table \ref{tab:mnist} shows quantitative results on the MNIST dataset. The accuracy of the tested models on domain $B$ is listed, which is artificially limited to $b$ samples (in brackets the share of $b$ w.r.t. the whole dataset is given). $T1$ and $T2$ denote the application of our framework while pre-training $\mathbf{T}$ with the identity and a $180^\circ$ rotation matrix, respectively. We limit the space of possible transformations to euclidean transformations (rotations + translations). Further limiting or relaxing this assumption slightly increases or decreases performance. $\mathbf{C}$ is a simple CNN consisting of 2 convolutional and one fully-connected layer. The used train mode is 0, setting it to 1 does not improve performance. Additionally, we show results of $\mathbf{M}$ trained on $A$ and applied to $B$ ($A$ on $B$), as well as of training all of $\mathbf{M}$ on the full dataset $B$ ($B$ on $B$) to give a theoretical upper bound of reachable performance. 

The drastic drop in accuracy when switching domains shows the need for transfer learning techniques.
\begin{table}[!b]
\caption{Results on the MNIST dataset.}
\centering
\label{tab:mnist}
\begin{tabular}{|c||c|c|}
 \hline
$b$ & Model  & Accuracy \\
 \hline
100 (0.14)  & Fine-tune & 0.738  \\
& CORAL & 0.5 \\

& pix2pix & 0.375 \\
& Imp & 0.313 \\

 & Ours - $T1$ & 0.766 \\
 & Ours - $T2$ & \textbf{0.912} \\
\hline
1000 (1.43) & Fine-tune & 0.908 \\
& CORAL & 0.852\\

& pix2pix & 0.898 \\
& Imp & 0.711 \\

 & Ours - $T1$ & 0.901 \\
 & Ours - $T2$ & \textbf{0.947} \\
\hline
2000  (2.86) & Fine-tune & \textbf{0.972} \\
& CORAL & 0.935\\
& pix2pix & 0.859\\ 
& Imp & 0.846 \\

 & Ours - $T1$ & 0.966 \\
 & Ours - $T2$ & 0.969 \\
\hline
70000 (100) & $A$ on $B$ & 0.153\\
& $B$ on $B$ & 0.980\\
\hline
\end{tabular}
\end{table}
Our framework performs at least as well or very comparably to fine-tuning. Incorporating domain knowledge ($T2$) greatly improves performance, already or especially for small $b$ a very high classification accuracy is reached. 
Fine-tuning and our variant $T1$ are close together, since learning to align images via pixel-loss is difficult \cite{perp, Mathieu2015DeepMV}. Further, on this limited dataset the models already achieve near-optimal performance.
Note that for images, since we use the same module, but also in general our approach with train mode 1, combining classification and pixel-loss, is very similar to STNs. However, since train mode 1 yields no performance increase, this shows that also STNs are not more effective in scenarios with limited training data.

As train mode 1 shows no advantages over train mode 0 in this example, Step 2 of our training procedure is of no big use, and thus results of Mode 0 and 2 are omitted here. 
For this problem, the original pix2pix architecture is used. This and baseline Imp both exhibit problems  dealing with this task: Although they perform well for local image transformations (e.g., changing intensities and colors of pixels), they fail for global ones, such as rotations. This shows the need for methods such as STNs or our framework. In addition, it shows the advantage of explicit transformations as calculated by our framework: Instead of hallucinating corresponding samples of the other domain, possibly worsening classification performance, among others due to blurring, transformation matrices offer invertible mappings, and can be initialized to meaningful values, or the identify function, at last. Performance of CORAL increases with growing $b$, but still does not reach that of fine-tuning or our models.

\subsection{Toy Sequence}
As this problem serves as preparation and motivation for the real-world problem of predicting lane changes, we here only consider fine-tuning as baseline. For predicting lane changes, all baselines are evaluated.

Upon inspection of Table \ref{tab:toysequence} we find our earlier assumption confirmed: a model trained on more well-behaved data has problems dealing with noisy trajectories, \emph{Frequency} nearly doubles when switching from domain $A$ to $B$. $T1$ again denotes our framework, in which $\mathbf{C}$ is pre-trained with the identity matrix. $T2$, however, uses the following pre-training target: During follow periods, the target matrix is $\begin{pmatrix}0 & 0 \\0.5 & 1\end{pmatrix}$, and during lane changes the identity. Due to this, the Converter is trained to output 0.5 (i.e. projecting the car onto the middle of the road) when its belief is lane following, and otherwise leaves the expected lane change unchanged. This causes a smoothing of trajectories during follow periods, making the trained model less subjective to noise, but still enables the correct detection of lane changes. 

\begin{figure}[!t]
\centering
\includegraphics[scale=0.4]{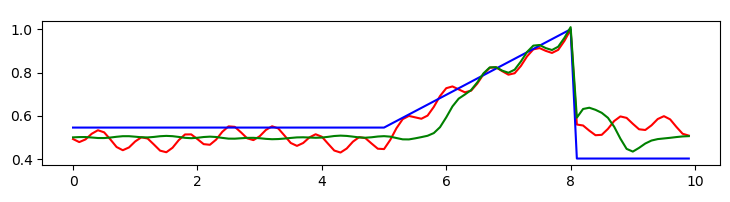}
\caption{Visualization of a simulated lane change to the left. The distance to the lane's center line is plotted on the y-axis, time in seconds on the x-axis. The ``noisy'' lane change from domain B is drawn in \textcolor{tred}{red}, a corresponding one from domain A in \textcolor{tblue}{blue} (for simplicity, just one of the $n$ is shown). The output of the Converter (with $T2$) is drawn in \textcolor{tgreen}{green}, and shows a very plausible converted lane change.}
\label{fig:toyseq}
\end{figure}

Results show the effect of this domain knowledge, while achieving similar \emph{Delay} and \emph{Miss} scores as other methods, $T2$ records by far the lowest \emph{Frequency}, even lower than the original model (more on this in the next section).
Depending on $b$, the ordering of fine-tuning and $T1$ changes, this dataset is too small and simplistic to learn meaningful transformations by itself - this is achieved for the next task. 
Figure \ref{fig:toyseq} shows the output trajectory of the Converter when applied to a lane change. The smoothing effect in the follow period is nicely visible, and the success of the conversion can be checked visually.
The hidden size of $\mathbf{C}$ is 16.
\begin{table}[!b]
\caption{Results of the toy sequence problem. Smaller values for \emph{Frequency}, \emph{Delay} and \emph{Miss} are better, larger ones for \emph{Score}.}
\vskip 0.15in
\centering
\label{tab:toysequence}
\begin{tabular}{|c||c|c|c|c|c|}
 \hline
$b$ & Model  & Frequency & Delay & Miss & Score \\
 \hline
 100  & Fine-tune & 3.196 &	0.945 &	0.121 & 0.186 \\
(1)  & $T1$ & 2.849 &	\textbf{0.897} &	\textbf{0.112} & \textbf{0.275} \\
 & $T2$ & \textbf{2.724} & 1.077	& 0.153 & 0.262 \\
 \hline
500    & Fine-tune & 3.456 &	0.888 &	\textbf{0.113} & 0.137 \\
 (5) & $T1$ & 3.156 &	\textbf{0.878} &	0.117 & 0.203 \\
 & $T2$ & \textbf{1.410} &	1.140 &	0.140 & \textbf{0.565} \\
 \hline
 2000 & Fine-tune & 2.940 &	\textbf{0.903} &	0.106 & 0.257 \\
 (20)  & $T1$ & 3.871 &	0.931 &	0.120 & 0.034 \\
 & $T2$ & \textbf{1.300} &	0.992 &	\textbf{0.104} & \textbf{0.625} \\
\hline
 10000  & $A$ on $B$ & 4.085 & 0.851 & 0.108 & - \\
 (100) & $B$ on $B$ & 2.129 & 0.542 & 0.062 & - \\
\hline
\end{tabular}
\end{table}

\subsection{Lane Change Prediction}
Table \ref{tab:lcp2} shows results of all analyzed models when considering the 2 features $m$ and $v$. A hidden size of 32 is used for $\mathbf{C}$. Results for experiments with more features are similar, especially regarding the ranking of algorithms. It can be noted though, that more features initially cause a greater performance drop when changing domains, as the learned knowledge is more specific, but perform better after the application of transfer learning techniques - after retraining the additional information is helpful. Further, fine-tuning performs worse, as the domain gaps get wider, and training the full model on $B$ better, especially reducing \emph{Frequency}, again due to the amount of available information (compare \cite{lcp} for more results).

\begin{figure*}[b]
\centering
\includegraphics[scale=0.42]{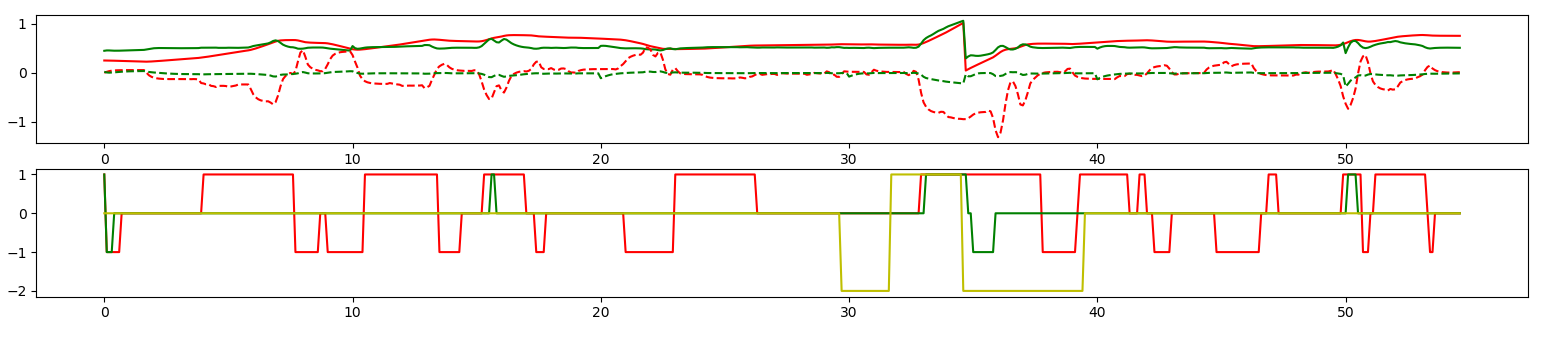}
\caption{Visualization of a lane change to the left, on the x-axis time in seconds is plotted. The top plot shows $m$, once the raw data from domain $B$ (\textcolor{tred}{red}) and once the Converter's output (\textcolor{tgreen}{green}). Same holds for $v$, this is indicated by the dashed lines. In the bottom plot the labels of the corresponding time steps are shown: The ground truth label is drawn in \textcolor{tyellow}{yellow}, the prediction of the fine-tuning approach in \textcolor{tred}{red}, the output of our model ($T2$) in \textcolor{tgreen}{green}. 1 / -1 denote lane changes to the left / right, 0 follow periods and -2 ignore labels, which are inserted between follow and lane change labels and after lane changes, to give the models time to reset. The effects of the Converter can clearly be seen, smoothing out fluctuations and scaling down extreme values of the input features, especially during follow periods. Our model outperforms standard fine-tuning, having much less false predictions while virtually predicting identically during lane changes.}
\label{fig:lcp}
\end{figure*}

The pre-training targets in Step 1 are the same as in the previous section (for follow periods the target values of $m$ and $v$ are neutral, i.e. 0.5 and 0).
For this task, our proposed models outperform all others, both with $T1$ and $T2$, in terms of \emph{Frequency} and total score. We note, that also $T1$ works well in this context, the Converter is able to gain enough knowledge to find meaningful transformations by itself. Furthermore, also for this task domain knowledge helps improving performance. Our models even perform ``better'' in terms of \emph{Frequency} than the full model trained on $B$. This can be explained by the fact that training is done by just using cross-entropy loss, and not by directly optimizing for any higher-level metrics. The found solution thus is a local optimum w.r.t. to this loss, resulting in a well-rounded solution. This shows other exciting properties of our framework: Potentially, it is not only applicable to transfer learning tasks in a strict sense, but also to tune and alter existing models in a desired way. Depending on pre-training, we could nudge models to prioritize completely different aspects.

When comparing our model to the baselines, we see a bigger difference for pretraining target $T1$. This makes sense, as all models profit from prior information ($T2$) and improve, and with less information given the selected correspondence pairs and the resulting correspondence loss is more  valuable. 

Thus, in this scenario the correspondence pairs and linked correspondence loss does help improve performance.
In particular, our full model outperforms all ablation studies (Mode 0 and Mode 2), and all other baselines. 

Figure \ref{fig:lcp} shows a sample sequence of aligned lane changes, plotting $m$ and $v$ as well as predictions and ground truth labels, comparing our model to fine-tuning only.

CORAL yields somewhat acceptable results, but drops behind fine-tuning. One reason could be the small size of the covariance matrix, and that for these high-level sequential, non-image inputs, domain differences express differently than via pure second-order statistics. Our used pix2pix adaption for this problem consists of simple, one-layered RNNs as generator (hidden size 32) and discriminator (hidden size 8). Arguably, this could be improved, but we found adversarial training on trajectories to be tedious and non-promising. Our implicit variant fares better, matching our full model with $T1$. This shows, on the one hand, that such a domain adaptation can be done implicitly, as well. On the other hand this proves, that we do not loose expressiveness by using transformation matrices. We would like to remind the reader of the advantages of such a scheme as described in the introduction, such as better verifyability, and the possibility of initializing $\mathbf{T}$ with domain knowledge: No method can compare with $T2$, especially for smaller $b$.



\begin{table}[!t]
\caption{Results of the lane change prediction problem. Smaller values for \emph{Frequency}, \emph{Delay} and \emph{Miss} are better, larger ones for \emph{Score}.}
\centering
\label{tab:lcp2}
\begin{tabular}{|c||c|c|c|c|c|}
 \hline
$b$ & Model  & Frequency & Delay & Miss & Score \\
 \hline
100 & Fine-tune & 7.344 & 0.612 & 0.008 & 0.352\\
 & CORAL & 7.996 & \textbf{0.601} & 0.011 & 0.291 \\
 & pix2pix & 5.399 & 0.837 & 0.039 & 0.439\\
  & Imp & 4.363 & 0.837 & 0.006 & 0.592 \\
(1.8) & Mode 0 - $T1$ & 7.48 & 0.637 & 0.008 & 0.339\\
 & Mode 0 - $T2$ & 5.477 & 0.738 & 0.008 & 0.502\\
 & Mode 2 - $T1$ & 5.835 & 0.699 & \textbf{0.005} & 0.48\\
 & Mode 2 - $T2$ & 5.243 & 0.833 & 0.011 & 0.51\\

 & Ours - $T1$ & 4.573 & 0.797 & 0.006 & 0.578\\
 & Ours - $T2$ & \textbf{3.373} & 0.956 & \textbf{0.005} & \textbf{0.672}\\
\hline
500 & Fine-tune & 7.744 & \textbf{0.544} & 0.01 & 0.319\\
 & CORAL & 8.336 & 0.547 & 0.011 & 0.265 \\
  & pix2pix & 6.003 & 0.842 & 0.071 & 0.321 \\
   & Imp & 4.476 & 0.835 & 0.010 & 0.576 \\
(9.1) & Mode 0 - T1 & 7.93 & 0.559 & 0.01 & 0.302\\
 & Mode 0 - $T2$ & 5.138 & 0.704 & 0.006 & 0.536\\
 & Mode 2 - $T1$ & 5.635 & 0.668 & 0.003 & 0.502\\
 & Mode 2 - $T2$ & 5.535 & 0.658 & 0.003 & 0.511\\

 & Ours - $T1$ & 4.848 & 0.759 & \textbf{0.002} & 0.565\\
 & Ours - $T2$ & \textbf{3.524} & 0.887 & \textbf{0.002} & \textbf{0.669}\\
\hline
1000 & Fine-tune & 6.551 & 0.620 & 0.005 & 0.424\\
 & CORAL & 6.933 & \textbf{0.608} & 0.010 & 0.383 \\
  & pix2pix & 2.339 & 1.157 & 0.144 & 0.460 \\
   & Imp & 3.803 & 0.800 & 0.006 & 0.641 \\
(18.2) & Mode 0 - $T1$ & 6.629 & 0.617 & 0.005 & 0.418\\
 & Mode 0 - $T2$ & 5.223 & 0.706 & 0.005 & 0.531\\
 & Mode 2 - $T1$ & 5.058 & 0.686 & \textbf{0.002} & 0.552\\
 & Mode 2 - $T2$ & 4.976 & 0.702 & 0.003 & 0.556\\

 & Ours - $T1$ & 3.594 & 0.901 & 0.01 & 0.646\\
 & Ours - $T2$ & \textbf{3.241} & 0.888 & 0.003 & \textbf{0.691}\\
\hline
5500 & $A$ on $B$ & 11.672 & 0.347 & 0.010 & -\\
(100) & $B$ on $B$ & 4.732 & 0.698 & 0.005 & -\\
\hline
\end{tabular}
\end{table}

\section{Conclusion}
We have proposed a general framework for transfer learning, applying it to image and sequential data, and performing better than existing methods. 
Primarily, we analyze the problem of predicting lane changes, for which we switch domains from proprietary fleet data to a public dataset. While doing so, our method shows a measurable smoothing of otherwise too sensitive prediction results, significantly reducing especially the false positive rate.
Using an explicit transformation matrix comes with benefits for many applications, such as the explainability of the method and the possibility to integrate often existing domain knowledge.



\bibliographystyle{IEEEtran}
\bibliography{output}

\end{document}